\documentclass[%
 reprint,
 amsmath,amssymb,
 aps,
]{revtex4-2}

\usepackage{graphicx}
\usepackage{dcolumn}
\usepackage{bm}
\usepackage{comment}
\usepackage{xcolor}
\usepackage{hyperref}

\def\ie{{\frenchspacing\it i.e.}}

\def\A{\textbf{A}}

\def\b{\textbf{b}}
\def\r{\textbf{r}}
\def\x{\textbf{x}}
\def\y{\textbf{y}}

\def\spose#1{\hbox to 0pt{#1\hss}}
\def\simlt{\mathrel{\spose{\lower 3pt\hbox{$\mathchar"218$}}
     \raise 2.0pt\hbox{$\mathchar"13C$}}}
\def\simgt{\mathrel{\spose{\lower 3pt\hbox{$\mathchar"218$}}
     \raise 2.0pt\hbox{$\mathchar"13E$}}}
\def\simpropto{\mathrel{\spose{\lower 3pt\hbox{$\mathchar"218$}}
     \raise 2.0pt\hbox{$\propto$}}}

\newcommand{\centered}[1]{\begin{tabular}{l} #1 \end{tabular}}

\begin{document}

\preprint{APS/123-QED}

\title{GenEFT: Understanding Statics and Dynamics of\\ Model Generalization via Physics-Inspired Effective Theory}

\author{David D. Baek}
 \email{dbaek@mit.edu}
\author{Ziming Liu}%
\author{Max Tegmark}%
 \email{tegmark@mit.edu}
\affiliation{%
 Massachusetts Institute of Technology, Cambridge, Massachusetts 02139, USA 
}%


\date{\today}

\begin{abstract}
We present GenEFT: an effective theory framework for shedding light on the statics and dynamics of neural network generalization, and illustrate it with graph learning examples.
We first investigate the generalization phase transition as data size increases, comparing experimental results with information-theory-based approximations. We then introduce an effective theory for the dynamics of representation learning, where latent-space representations are modeled as interacting particles (``repons''), and find that it explains our experimentally observed phase transition between 
generalization and overfitting as encoder and decoder learning rates are scanned.
This highlights the potential of physics-inspired effective theories for bridging the gap between theoretical predictions and practice in machine learning.
\end{abstract}

\maketitle


\section{Introduction}

A key feature of intelligence is the ability to generalize -- applying knowledge learned in one context to new, unseen situations. This hallmark of human intelligence is mirrored in machine learning, where the objective is to develop models that generalize beyond the training data to make accurate predictions on new, unseen inputs. Generalization is what separates a useful machine learning model from an overfitted one~\cite{bejani2021systematic,salman2019overfitting,srivastava2014dropout}, making it a crucial property for ensuring reliable performance in real-world scenarios. Without proper generalization, a model may appear to perform well on training data but fail to make correct predictions when faced with new data, rendering it ineffective for practical applications.

However, achieving generalization is not guaranteed and depends on various hyperparameters, such as the amount of training data, the learning rate, model complexity, and regularization techniques. If these parameters are not carefully set, the model may overfit -- learning spurious correlations that do not generalize beyond the training set. For instance, a model trained with too little data might not capture essential patterns, while an excessively high learning rate could prevent proper convergence, leading to unstable or suboptimal solutions. Conversely, an overly small learning rate might cause the model to get stuck in a narrow local minimum, failing to explore a broader region of the loss landscape that could yield better generalization. Understanding how these factors influence generalization is critical for designing effective machine learning systems.

Early statistical and measure-theoretic approaches~\cite{kawaguchi2018generalization,mukherjee2006learning,xu2012robustness,vapnik1999overview} to understanding generalization primarily focused on developing various training and regularization algorithms to minimize the generalization error -- the gap between empirical loss and theoretical loss. Although these methods do improve generalization, they do not provide insights into how different hyperparameters affect generalization. More recent works have directly tackled this problem, including studies of various learning phases (i.e., memorization, generalization, and grokking) as a function of the learning rate~\cite{liu2021towards,ye2021towards,liu2022omnigrok,power2022grokking,lewkowycz2020large}, investigating how the structural conditions of a model affect the representations in a Restricted Boltzmann Machine~\cite{tubiana2017emergence}, studying learning dynamics of neural networks~\cite{lyle2022learning, hu2020surprising, fu2024learning}, and studying various data augmentation strategies to improve generalization~\cite{barbiero2020modeling,jha2020does,volpi2018generalizing,hernandez2018data,hansen2021generalization}. While these studies are highly insightful for enhancing our understanding of generalization, (a) their analyses are mostly limited to specific model architectures or problems, and (b) they have yet to offer a simple, guiding theoretical framework that can estimate key training hyperparameters -- such as the learning rate and the required amount of training data -- which are crucial for practical applications.

\begin{figure*}[t]
\begin{center}
\includegraphics[width=.58\linewidth]{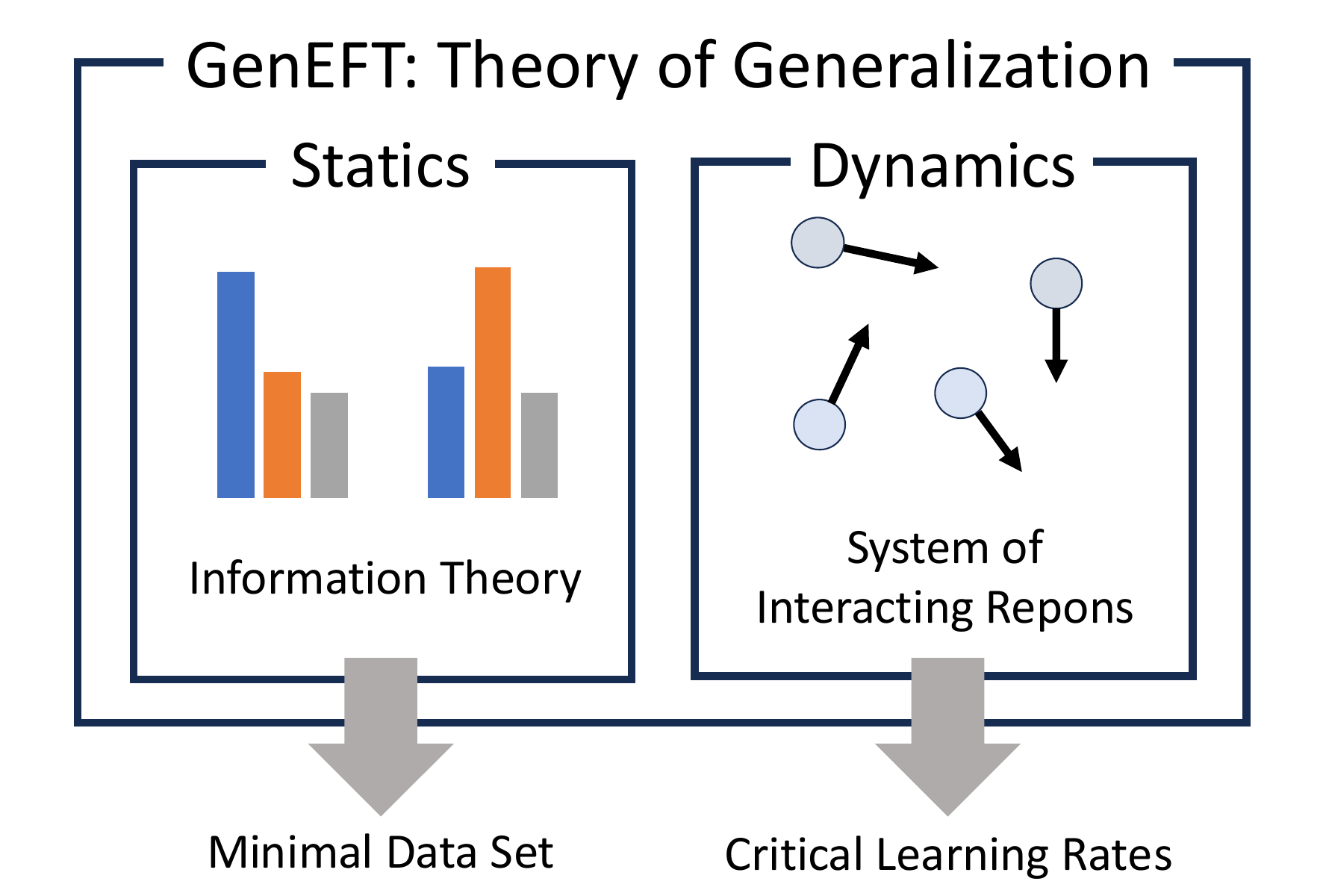}
\end{center}
\caption{\textcolor{black}{Summary of GenEFT: Physics-inspired effective theory for understanding statics and dynamics of model generalization.}
}
\label{framework-diag}
\end{figure*}

In this paper, we introduce GenEFT, a simple physics-inspired approximation framework that not only provides useful estimates of key training hyperparameters, thereby bridging the gap between theoretical generalization bounds and practical machine learning training strategies, but also advances and unifies our understanding of generalization. We propose two closely intertwined, solvable models: one model, addressing \emph{statics} of generalization, examines how data shape the final generalization outcome, while the other, focusing on \emph{dynamics} of generalization, explores how learning rates govern the training process and, ultimately, the generalization behavior. We believe that physics-inspired approaches are especially well-suited for this task because physics provides powerful tools for developing practical approximations to model complex systems.

We note that generalization is not solely a property of the dataset or the training process--it emerges from the interaction between the two. A well--curated dataset alone does not ensure generalization if the optimization dynamics are flawed, just as an optimal learning rate cannot compensate for inadequate or biased training data. Therefore, true generalization arises from a harmonic combination of these two factors, requiring a deeper understanding of both data distribution and learning dynamics. Hence, our effective theory addresses both parts: \emph{statics} part allows understanding how \emph{data} affect final generalization outcome, and \emph{dynamics} part allows understanding how hyperparameters such as learning rate affect the learning dynamics.

In \emph{statics} section, we present an information-theoretic approximation that describes the relationship between test accuracy and the amount of fraction of training data. For graph learning, full generalization only becomes possible after training on at least $b$ examples, where $b=\log_2 N$ is the number of bits required to specify which of $N$ possible graphs is present, and further delays are caused both by correlated data and by the {\it induction gap}, bits required to learn the inductive biases (graph properties) that enable generalization.

In \emph{dynamics} section, we present \emph{Interacting repon theory}, where the representations are modeled as interacting particles that we term ``repons'' and their dynamics are described as interacting system of particles. We show that by solving this model, one could predict how the  memorization/generalization phase transition depends on encoder and decoder learning rates.

In summary, we conclude that (a) static/steady state properties of generalization, such as the critical training fraction, could be derived from information-theoretic properties of the input, and (b) dynamic properties of generalization, such as critical learning rate, could be derived from studying the system of interacting repons, particles whose locations are given by their embedding representations. This framework is summarized in Figure \ref{framework-diag}.

The rest of this paper is organized as follows: In Section \ref{prob_form}, we formally describe our problem settings. In the following two sections, Section \ref{critical_train_fraction}--\ref{critical_lr}, we present an effective theory for understanding statics and dynamics of generalization using a combination of theoretical frameworks and empirical evidence. We summarize our conclusions in Section~\ref{conclusion}.

\section{Problem Formulation}
\label{prob_form}

In order to explore generalization properties, we examine a knowledge graph learning problem using an autoencoder. Our setting is motivated by the fact that a lot of real-world problems could be mapped into a knowledge graph learning problem since they inherently involve entities with complex interrelationships. For example, WordNet, a lexical database of the English language, can be viewed as a knowledge graph where words are nodes and their semantic relationships (such as synonyms, hypernyms, and hyponyms) are edges.

Regarding the choice of architecture, the autoencoder architecture is chosen as a representative considering the fact that autoencoders are a widely adopted training paradigm in machine learning. Such widespread use is reflected in the design of many modern language models, such as GPT-4~\cite{achiam2023gpt}, LLaMA~\cite{dubey2024llama}, and Gemma~\cite{team2024gemma}, which leverage encoder-decoder architectures.

We formalize our knowledge graph learning problem as follows: Consider an arbitrary binary relation $R$ of elements $x_1,...,x_n$, for example ``greater-than'' or ``equal modulo 3''. 
This can be equivalently viewed as a directed graph of order $n$, and is specified by
the $n\times n$ matrix $\mathbf{R}_{ij}\equiv R(x_i,x_j)\in\{0,1\}$.
Our machine-learning task is to predict the probability that $R(x_i,x_j)=1$ by training on a random data subset. 

\textbf{Model Architecture:} We use an autoencoder (encoder-decoder) architecture where the encoder
maps each input $x$ into a latent space representation $\mathbf{E}_{x}\in\mathbb{R}^d$
after which an MLP decoder concatenates two embeddings $[\mathbf{E}_{x_1}, \mathbf{E}_{x_2}]$
and maps it into the probability estimate. We use $\tanh$ activation in any hidden layers followed by 
a sigmoid function to obtain the probability and use cross-entropy loss. The learnable parameters in the model are the embedding vectors as well as the weights and biases in the decoder MLP.

\textbf{Datasets} The relations $R$ that were used in our experiments are equal modulo 3, equal modulo 5, greater-than, and complete bipartite graph, whose formal definitions are summarized in Table \ref{relation-list}. These relations were chosen to provide a broad diversity of relation properties. For instance, equal modulo 3 and equal modulo 5 are equivalence classes (symmetric, reflexive, and transitive), greater-than is a total ordering (antisymmetric and transitive), while the complete bipartite graph is symmetric and anti-transitive. We will see that our effective theory is applicable to these general relations regardless of their detailed properties. We used $n=30$ elements
for our experiments.  For each relation, we thus have $n^2=900$ data samples $\mathbf{R}_{ij}$. The training data was randomly sampled from this set, and the remaining samples were used for validation (testing).

\begin{table*}[ht]
\caption{Relations used in the experiments. $S$ is one of two disjoint sets in a bipartite graph.
}
\label{relation-list}
\begin{center}
\begin{tabular}{ll}
\multicolumn{1}{c}{\bf Relation}  &\multicolumn{1}{c}{\bf Description}
\\ \hline \\
Modulo $3$         & $(x,y) \in R$ iff $x \equiv y \;(\textrm{mod}\; 3)$ \\
Modulo $5$         & $(x,y) \in R$ iff $x \equiv y \;(\textrm{mod}\; 5)$ \\
Greater-than       & $(x,y) \in R$ iff $x < y$ \\
Complete Bipartite Graph $G=(S,\overline{S})$    & $(x,y) \in R$ iff ($x \in S, y \notin S$) or ($x \notin S$, $y \in S$) \\
\end{tabular}
\end{center}
\end{table*}

\begin{table*}[ht]
\caption{Bits required to describe relations with various properties.
}
\label{relation-dof-table}
\begin{center}
\begin{tabular}{ll}
\multicolumn{1}{c}{\bf Relation}  &\multicolumn{1}{c}{\bf Description length $b$ [bits]}
\\ \hline \\
Generic & $n^2$ \\
Symmetric &  $n(n+1)/2$ \\
Antisymmetric &  $n(n-1)/2$ \\
Reflexive &  $n(n-1)$ \\
Transitive & $\approx\frac{n^2}{ (\langle k \rangle -1)!}$, $\langle k \rangle$ is the average length of a transitive chain \\
Equivalence Relation with $k$ Classes         & $\approx n\log_2 k$ \\
Total Ordering       & $\log_2 n!\approx n \textrm{log}_2{n\over e}$ \\
Complete Bipartite Graph    & $n$ \\
Incomplete Bipartite Graph $G=(S_1,S_2)$    & $|S_1||S_2|$ \\
Tree Graph &  $\approx n \textrm{log}_2 \langle d \rangle$ where $\langle d \rangle$ is the average node depth
\end{tabular}
\end{center}
\end{table*}

\section{Minimal Data Amount for Generalization}
\label{critical_train_fraction}

In this section, we develop an effective theory to understand \emph{static} properties of the trained neural network, with a particular focus on their generalization capability. Specifically, we identify the existence of a set of \emph{critical} information (\emph{bits}) that must be learned during training to enable a model to generalize effectively. These critical bits correspond to relational properties, such as transitivity, symmetricity, and reflexivity, in the context of knowledge graph learning. Building on this insight, we formulate an effective theory from an information-theoretic perspective to quantify the data necessary for learning these critical bits.

\textbf{Effective Theory: Information-theoretic Approach} Let $b$ represent the description length of the relation, which refers to the minimum number of independent bits required to fully represent the relation. This length corresponds to the smallest amount of information (in terms of bits) necessary to encode all the relevant details of the relation without redundancy or loss of essential information. For graphs sampled uniformly from a class with certain properties (transitivity and symmetry, say), 
$b = \log_2 C$, where $C$ is the total number of graphs satisfying these  properties. For instance, there are $n!$ total orderings, giving $b\approx n\log_2(n/2)$ using Stirling's approximation.

An arbitrary real-world graph will likely have properties that go beyond named properties such as transitivity and symmetry. The total number of graphs with the same structure as a graph $G$ is given by $\frac{n!}{|\operatorname{Aut}(G)|}$, where ${\rm Aut}(G)$ is the automorphism group of $G$ \cite{mathon1979note}. The relation's description length is thus $b = \log_2 \frac{n!}{|\operatorname{Aut}(G)|}$. For a graph with total ordering, $|\operatorname{Aut}(G)| = 1$ so $b = \log_2 \frac{n!}{|\operatorname{Aut}(G)|} = \log_2 n!$, which matches the corresponding value in Table \ref{relation-dof-table}. For a complete bipartite graph $K_{a, c}$ ($a \ne c$, $a + c = n$), we have $|\operatorname{Aut}(G)| = a!c!$. In this case, $b = \log_2 \frac{n!}{a!c!} = \log_2 \binom{n}{a} \approx kn$, which matches the corresponding value in Table \ref{relation-dof-table} up to a constant factor. We can compute $|\operatorname{Aut}(G)|$ for arbitrary sparse graphs (e.g., knowledge graphs) using open-source software such as SAUCY \cite{darga2008faster}, which takes seconds to process graphs with millions of nodes.

\begin{figure}[tb]
\begin{center}
\includegraphics[width=.9\linewidth]{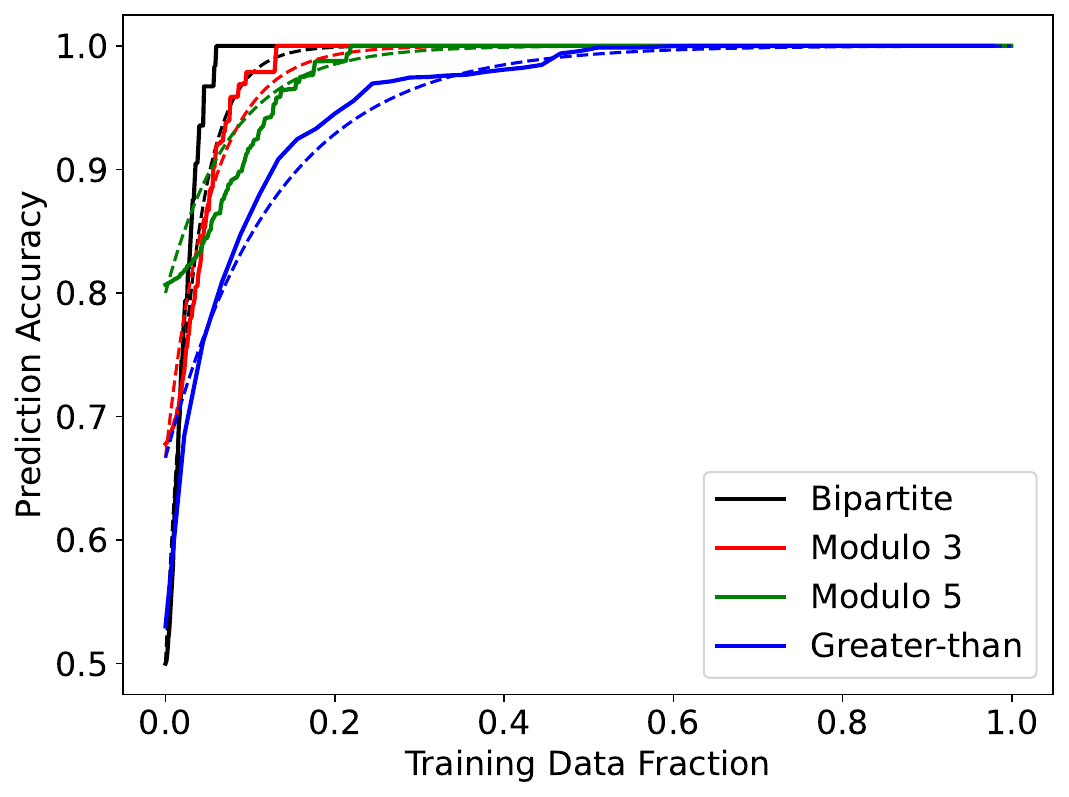}
\end{center}
\caption{Plot of numerical experiments (solid line) and analytic formula in Eq. \ref{eff:formula} (dotted line) for the theoretical upper bound of prediction accuracy $f_{\rm UB}$ as a function of training data fraction.}
\label{critical-fraction-theory}
\end{figure}

\begin{figure*}[tb]
\begin{center}
\includegraphics[width=.84\linewidth]{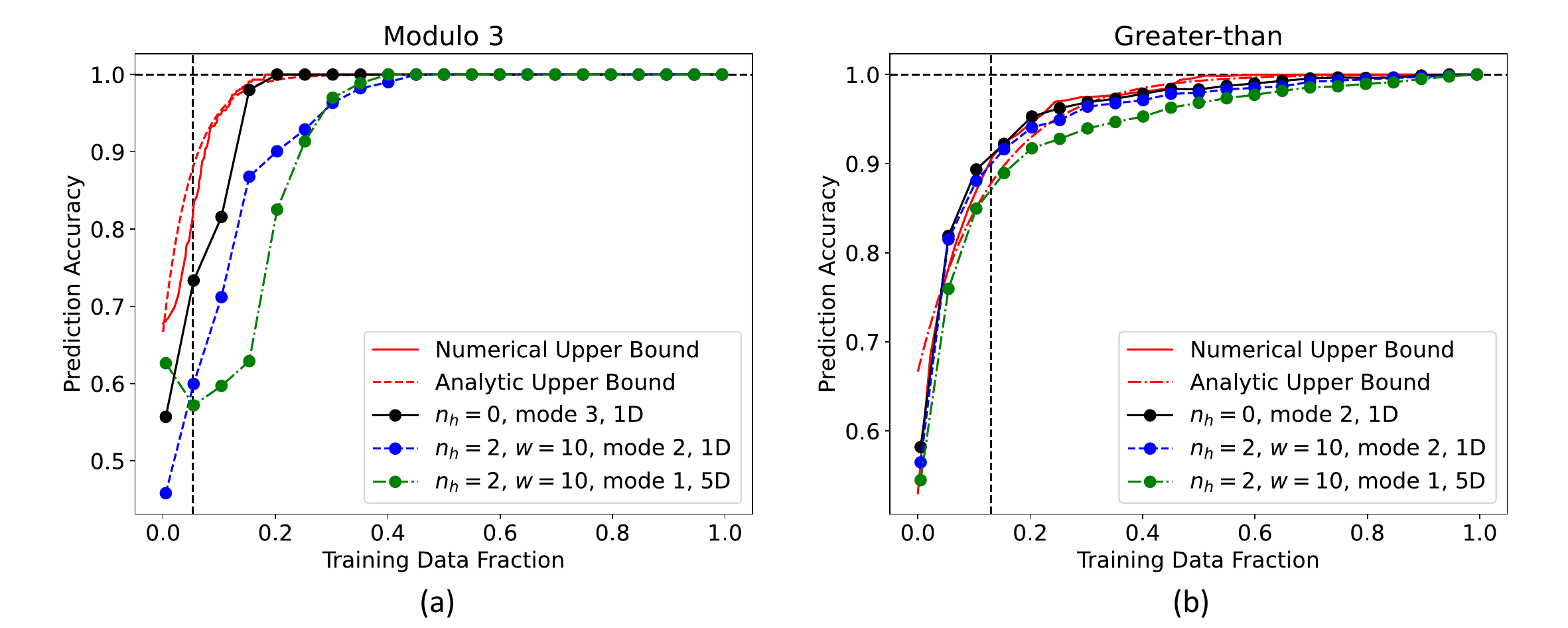}
\end{center}
\caption{Prediction accuracy vs.~training data fraction for learning the relations (a) equivalent modulo 3, and (b) greater-than. We see that
applying inductive bias to the decoder MLP architecture (making it simpler while still capable to succeed) improves the performance 
(reduces the ``induction gap'' gap relative to the theoretical upper bound). In the legend, $n_h$ is the number of hidden layers, $w$ is the width of the hidden layers, mode indicates how the two embeddings are combined, and the dimension is that of the embedding space. 
Vertical lines show the approximate transition scale $b/n^2$.}
\label{critical-fraction-exp}
\end{figure*}

To fully generalize and figure out precisely which graph in its class we are dealing with, we therefore need at least $b$ training data samples \emph{on expectation}, because the expectation value of how much information we learn from each sample is at most one bit.
As seen in Figures~\ref{critical-fraction-theory} and~\ref{critical-fraction-exp}, 
$b$ indeed sets the approximate scale for how many training data samples are needed to generalize, but to generalize perfectly, we need more than this strict minimum for two separate reasons. 
\begin{enumerate}
\item $k$ data samples may provide less than $k$ bits of information because (a) they are unbalanced (giving 0 and 1 with unequal probability) and (b) they are not independent, with some inferable from others: for example, the data point $ R(x_1,x_2)=1$ for a symmetric relation implies that $R(x_2,x_1)=1$.
\item We lack {\it a priori} knowledge of the {\it inductive bias} (what specific properties the graph has), and need to expend some training data information to learn this -- we call these extra bits the {\it inductive gap}. 
\end{enumerate}

Figure~\ref{critical-fraction-theory} shows the theoretical maximum accuracy attainable, computed in two steps. 
(a) First, we count the total number of directly inferable data pairs using \emph{a priori} knowledge of the relation -- for example, symmetry reflexivity, and transitivity for equivalence relations -- given the randomly sampled set of training data pairs.
(b) Second, we compute the probability $p_*$ of there being an edge (\ie \; $ P(R_{i, j} = 1)$) for each case that is not directly inferable from the training data. We do this via Monte Carlo simulation, considering all possible equivalence class memberships (or all possible orderings for the greater-than relation) that are compatible with the training data. The plotted numerical prediction accuracy is a sum of 1 for each inferred relation and ${\rm max}(p_*,1-p_*)$ for each non-inferred one, all divided by the total number of examples.

We also provide a simple but useful approximation for how much training data is needed to read a testing accuracy $\alpha$ (say 90\%).
Let the $b$ bits that describe the relation be $x_1, x_2, \cdots x_b$, and the number of training data samples is $m$. In the crude approximation that each data sample provides one bit worth of information, this is essentially the problem of putting $m$ distinguishable balls into $b$ different buckets, and we want at least a fraction $\alpha$ of the buckets to be non-empty. The ball that is thrown into an already non-empty bucket does not provide extra information about the relation and therefore, does not affect the prediction accuracy. We use the indicator variable $Z_i$: $Z_i=1$ iff the bucket $x_i$ is empty, $Z_i=0$ otherwise. Using the linearity of expectations,
\begin{align}
    E\left(\sum_i Z_i\right) = \sum_i E(Z_i) = b\left(1-\frac{1}{b}\right)^m \; \\
    \therefore E\left(\sum_i (1-Z_i) \right) = b\left[1- \left(1-\frac{1}{b}\right)^m\right].
\end{align}
The fraction of inferable data is hence \begin{equation}
    f(m) \approx \frac{1}{b} E\left(\sum_i (1-Z_i) \right) = \left[1- \left(1-\frac{1}{b}\right)^m\right]. \label{eff:formula}
\end{equation}
The critical training data fraction $p_c$ is given by $f(N p_c)=\alpha$, 
where $N=n^2$ is the total number of training samples. This gives the following formula for critical training data fraction:
\begin{equation}
    p_c = \frac{1}{N}\frac{\textrm{log}_2 (1-\alpha)}{\textrm{log}_2 \left(1-\frac{1}{b}\right)},
\end{equation}
where the only dependence on the properties of the relation is through its description length $b$.
In addition to directly inferable samples using \emph{a priori} knowledge of the relation, one can also guess the answer and get it correct. Letting $p_*$ denote the probability of guessing correctly, the approximation Eq. (\ref{eff:formula})
gives an upper bound 
$f_{\rm acc}\le f_{\rm UB}$ for the prediction accuracy $f_{\rm acc}$, where
\begin{equation}
    f_{\rm UB}(m) = f(m) + \textrm{max}(p_*,1-p_*) (1-f(m)) \label{eff:form_guess}.
\end{equation}


For an equivalence relation with $k$ classes, $p_*=1/k$. For a complete bipartite graph, $p_*=1/2$, since there are only two clusters that each element can belong to. For the greater-than relation, $p_*\approx1/3$. In order to understand this, consider a unit module consisting of three symbols $\{x,y,z\}$, and we are given $x<y$. The only three possible orderings of the three symbols are: $x<y<z$, $x<z<y$, and $z<x<y$. Since each of these orderings are equally likely to be the actual ordering, the probability of correctly guessing the pair $(x,z)$ or $(y,z)$ is $p_*=1/3$. We also numerically validate this in Figure \ref{critical-fraction-theory}.

The dashed and solid curves in Figure~\ref{critical-fraction-theory} show rough agreement between this approximation and the more exact Monte Carlo calculation for the inferable data fraction. The details of Monte Carlo calculations are explained in Appendix \ref{app:mc-details}.
One difference is that the training data include a certain level of information also 
about the pairs that are not directly inferable, improving the aforementioned accuracy term $p_*$.
Another difference is that data samples on average contain less than one bit (because they are not independent),
and that some samples contain more information than others.
For example, consider the task of learning greater-than relation on the numbers $0$ to $9$. 
The sample $0<9$ never provides any extra information, whereas the sample $4<5$ could be powerful 
if we already know the order of numbers $0$ to $4$ -- the comparison of at least 4 additional pairs, $(0,5), (1,5), (2,5), (3,5)$ 
can then be inferred. Figure \ref{critical-fraction-theory} compares our numerical experiments (solid line) with our analytic approximation (dotted line, Eq. (\ref{eff:form_guess})).


\textbf{Neural Network Experiments} Figure \ref{critical-fraction-exp} shows the empirical results of prediction accuracy vs. training data fraction for our neural network experiments. For direct comparison with our theoretical curves, prediction accuracy was computed over the entire dataset, including both the training data and the testing data.

In Figure~\ref{critical-fraction-exp}, we see that the neural generalization curves roughly track our theoretical prediction, with a slightly delayed rise, presumably for the two aforementioned reasons. Note that this gap between theory and experiment is different from the grokking delay \cite{liu2021towards}, which refers to the delay between the rise of the training and test accuracy curves.
 
We term the architecture-dependent part of this delay the {\it inductive gap}.
Figure \ref{critical-fraction-exp} illustrates that the inductive gap decreases when it is simplified by including more inductive bias in the model. We will now aim to provide intuition for this finding.
 
We compared three modes for combining encoder outputs into decoder input:
\begin{enumerate}
\item The default mode, where the input to the decoder is the concatenation of two embeddings $[\mathbf{E}_{x_1},\mathbf{E}_{x_2}]$, 
\item The decoder's input is $(\mathbf{E}_{x_1}-\mathbf{E}_{x_2})$, and 
\item The decoder's input is $|\mathbf{E}_{x_1}-\mathbf{E}_{x_2}|^2$, element-wise squared.
\end{enumerate}

Our choice of these modes is motivated by the fact that certain mathematical relations, such as equal modulo 3 or greater-than, depend solely on the difference between two numbers. Therefore, using the difference between two embeddings as input to the decoder introduces the appropriate inductive bias.

For learning equivalence modulo 3, the ideal embedding would be to cluster
numbers into three distinct positions on a 1-dimensional line corresponding to the 3 equivalence classes.
Hence, one expects the neural network to learn that $P[R(x_1,x_2)=1]=1$ 
iff $\mathbf{E}_{x_1} = \mathbf{E}_{x_2}$. This is distinguished by mode 3 followed by a single ReLU$(0.1-x)$ activation, say.
This embedding and decoder automatically enforce the symmetry, transitivity, and reflexivity of this relation, enabling it to exploit these properties to generalize.
Indeed, Figure \ref{critical-fraction-exp}(a) shows that the inductive gap is the smallest for a network in mode 3 with 0 hidden layers.
In contrast, the more complex neural networks in the figure waste some of the training data just to learn this useful
inductive bias.

 For learning the greater-than relation, the ideal encoding would be to embed numbers on a 1-dimensional line in their correct order.
 Hence, one expects the neural network to learn $P[R(x_1,x_2)=1]=1$ iff $({E}_{x_1} -{E}_{x_2} ) > 0$. This is best learned in mode 2: 
 the relation holds if the decoder's input is positive -- which can be accomplished with a single ReLU activation. 
 Indeed, Figure \ref{critical-fraction-exp}(b) shows that the inductive gap is the smallest for a network in mode 2 with 0 hidden layers.
In summary, we have illustrated that applying correct inductive bias to the neural network's architecture reduces the inductive gap, thereby improving its generalization performance.

While the statics analysis has clarified how the intrinsic information content of the data sets a critical threshold for generalization, it does not capture the full temporal evolution of the learning process. In the next section, we shift our focus to dynamics by introducing the interacting repon theory. Here, representations are modeled as interacting particles--``repons''--whose behavior over time reveals how variations in encoder and decoder learning rates precipitate the transition from memorization to generalization. This dynamic perspective not only complements our information-theoretic insights but also provides a deeper understanding of how training hyperparameters drive the evolution of learned representations.

\section{Interacting Repon Theory: Critical Learning Rates for Generalization}
\label{critical_lr}

In this section, we present an effective theory to explain the \emph{dynamics} of neural network training. Notably, we identify a duality between the problem of interacting particle systems in physics and neural network training. Leveraging this analogy, we model the representation learning process as a many-body system evolution problem and predict key properties. 
Specifically, we interpret each representation vector as a ``repon'' particle, moving in response to interactions with other repons. We show that when the optimal representation necessitates these particles to form distinct clusters (e.g., for learning equivalence relations), the nature of their interactions dictates the outcome: attraction between same-cluster repons fosters generalization, while repulsion results in overfitting.

We begin by analyzing the optimal representation for classification problems in Theorem 1. Next, we explore the dynamics of representation learning in Theorems 2 and 3, with Theorem 3 specifically predicting the critical learning rates required for generalization.

\noindent \textbf{(Theorem 1)} Consider training an autoencoder for a classification problem, where the input consists of two nodes, and the output is 1 if the two input nodes belong to the same class, and 0 otherwise. For an injective decoder with zero training loss, any two nodes must be embedded at the same point in the embedding space if they have the same label, and at different points otherwise. In other words, representations must be grouped based on their class membership in classification problems.

\noindent \emph{Proof}: Let $\textrm{Dec}(E_x || E_y)$ denote the operation of the decoder in the autoencoder, which takes the concatenation of the embeddings $E_x$ and $E_y$ as input.

Suppose two nodes $i$ and $j$ belonging to different classes have the same representation, i.e., $E_i=E_j$. Now, consider a node $k$ belonging to the same class as $i$. Note that the decoder output is the probability that the two input nodes belong to the same class. Now, we have $Y_{i,k} \equiv \textrm{Dec}(E_i || E_k) = \textrm{Dec}(E_j || E_k) = Y_{j,k}$, which is in contradiction with the fact that $1= Y_{i,k} \neq Y_{j,k}=0$. For two nodes $i$ and $j$ in the same class, $Dec(E_i || E_k) = 1 = Dec(E_j || E_k)$ and the injectivity of the decoder implies $E_i=E_j$, completing the proof.

\noindent 

\noindent \textbf{(Theorem 2)} Consider an interacting system of two same-class repons. In the limit of infinite time ($t \rightarrow\infty$), the distance between repons always converges to zero or a positive constant.

\noindent \emph{Proof}: We consider what happens
when two embeddings  $\x_1$ and $\x_2$
 are sufficiently close that the decoder can be locally linearized as $D(\x)\approx \A \x+\b$. 
 If the two belong in the same cluster, so that the loss function is minimized for $D(\x_1)=D(\x_2)=\y$, say, then the training loss for this pair of samples can be approximated as
\begin{equation}\label{eq:loss}
    \ell = \frac{1}{2}(|\A\x_1+\b-\y|^2+|\A\x_2+\mathbf{b}-\mathbf{y}|^2).
\end{equation}
Both the decoder $(\A,\b)$ and the encoder (the repon positions $\x_1$, $\x_2$) are updated via a gradient flow, with learning rates $\eta_A$ and $\eta_x$, respectively:
\begin{equation}
    \frac{d\mathbf{A}}{dt} = -\eta_A\frac{\partial \ell}{\partial \mathbf{A}}, \frac{d\mathbf{b}}{dt} = -\eta_A\frac{\partial \ell}{\partial \mathbf{b}}, \frac{d\mathbf{x}_i}{dt} = -\eta_x\frac{\partial \ell}{\partial \mathbf{x}_i}\ (i=1,2).
\end{equation}
Substituting $\ell$ into the above equations simplifying as detailed in Appendix \ref{app:derivation_3}, we obtain
\begin{equation}\label{eq:gradient_flow}
    \begin{aligned}
    &\frac{d\A}{dt} = -2\eta_A\A\r\r^T, \; \frac{d\r}{dt} = -\eta_x \A^T\A\r,\\
    \end{aligned}
\end{equation}
where $\mathbf{r}\equiv(\mathbf{x}_1-\mathbf{x}_2)/2$ is half the repon separation. 

This repon pair therefore interacts via a quadratic potential energy $U(\mathbf{r})=\frac{1}{2}\mathbf{r}^T\A^T\A\mathbf{r}$, leading to a linear attractive force $F(\r)\propto -\A^T\A\r$
which is accurate whenever two repons are sufficiently close together.
In one embedding dimension, this is like Hookes' law for an elastic spring pulling the repons together, except that the spring is weakening over time.
Eq.~(\ref{eq:gradient_flow}) has simple asymptotics: (i) If $\mathbf{A}$ is constant or $\eta_A=0$, two repons will eventually collide, i.e., $r\to0$. (ii) In the opposite limit, if $\mathbf{r}$ is constant or $\eta_x=0$, $\A$ will converge to $\mathbf{0}$ while $r>0$. In the general case where $\eta_x,\eta_A>0$, Eq.~(\ref{eq:gradient_flow}) is a competition between which of $\A$ and $\r$ decays faster, whose outcome depends on  $\eta_A$ and $\eta_x$, as well as the initial values of $\mathbf{A}_0$ and $\mathbf{r}_0$.
Without loss of generality, we can rescale the embedding space so that $|\r_0|=1$.

{\bf General case}
It is easy to check Eq.~(\ref{eq:gradient_flow}) has a solution of the following form:\begin{equation}\label{eq:ansatz}
    \mathbf{A}(t) = a(t)\mathbf{A}_0 + b(t)\mathbf{A}_0\mathbf{r}_0\mathbf{r}_0^T,\quad 
    \mathbf{r}(t) = c(t)\mathbf{r}_0.
\end{equation}
Substituting this into Eq.~(\ref{eq:gradient_flow}) and defining 
$a_2(t)\equiv a(t)+b(t)$
gives the  differential equations 
\begin{equation}\label{eq:toy_ode}
    \begin{aligned}
    & \frac{da_2}{dt} = -2\eta_Ac^2a_2, \;\frac{dc}{dt} = -\eta_x a_2^2c,\\
    \end{aligned}
\end{equation}
whose fixed point $(a_2^*,c^*)$ defined by $da_2/dt=0,dc/dt=0$ requires $a_2^*=0$ or $c^*=0$. 
A repon collision happens if $c^*=0$. Interestingly, there is a conserved quantity
\begin{equation}
    C\equiv \frac{1}{2\eta_A}a_2^2 - \frac{1}{\eta_x}c^2,
\end{equation}
because $dC/dt=0$. This defines a hyperbola and the value $C$ is determined by initialization. If $C>0$, $c^*\to 0$ and $a_2^*\to\pm\sqrt{2\eta_AC}$, resulting in repon collision (generalization). If $C<0$, $c^*\to\pm\sqrt{-\eta_xC}$ and $a_2^*\to 0$, leading to no collision (memorization). Here, we note that memorization does not always equate to overfitting to training data. In generative modeling, for example, a model may memorize useful structural patterns or high-level abstractions from the training distribution without merely replicating training examples.

\begin{figure}
    \centering
    \includegraphics[width=0.6\linewidth]{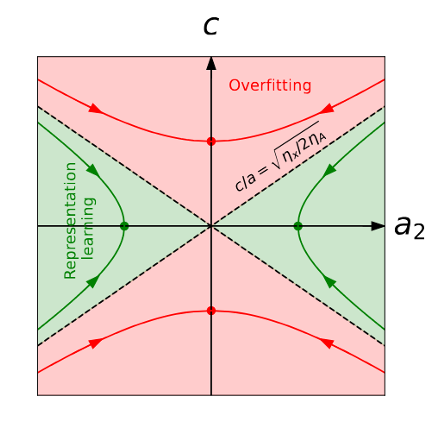}
    \caption{Phase diagram of interacting repon dynamics in classification problems. \textcolor{black}{The two axes are $a_2$ and $c$, which parameterize the decoder weights and the distance between two representations, respectively. Their formal definitions can be found in the main text. In this phase diagram, the green area represents initializations that lead to representation collision, thereby enabling generalizable representation learning, while the red area corresponds to regions that lead to overfitting or memorization.}}
    \label{fig:repon-ps}
\end{figure}

The mathematical form of the conserved quantity exhibits a striking similarity to that of classical harmonic oscillator problem in physics, where each term corresponds to quadratic kinetic and potential energy term. The dynamics of the system could hence be understood in $(a_2,c)$ phase space, where $a_2$ parameterizes decoder weights, and $c$ parameterizes representations, or more specifically the distance between two repons. The result of Theorem 2 implies that there are certain parts of the phase space (initializations) that are fundamentally forbidden to arrive at perfectly generalizable representation, which requires repon collision in classification problems. Figure \ref{fig:repon-ps} shows the phase diagram of interacting repon dynamics in classification problems, where the green area corresponds to initializations that could lead to repon collision, and hence generalizable representation learning.

\begin{figure*}[tb]
\begin{center}
\includegraphics[width=.99\linewidth]{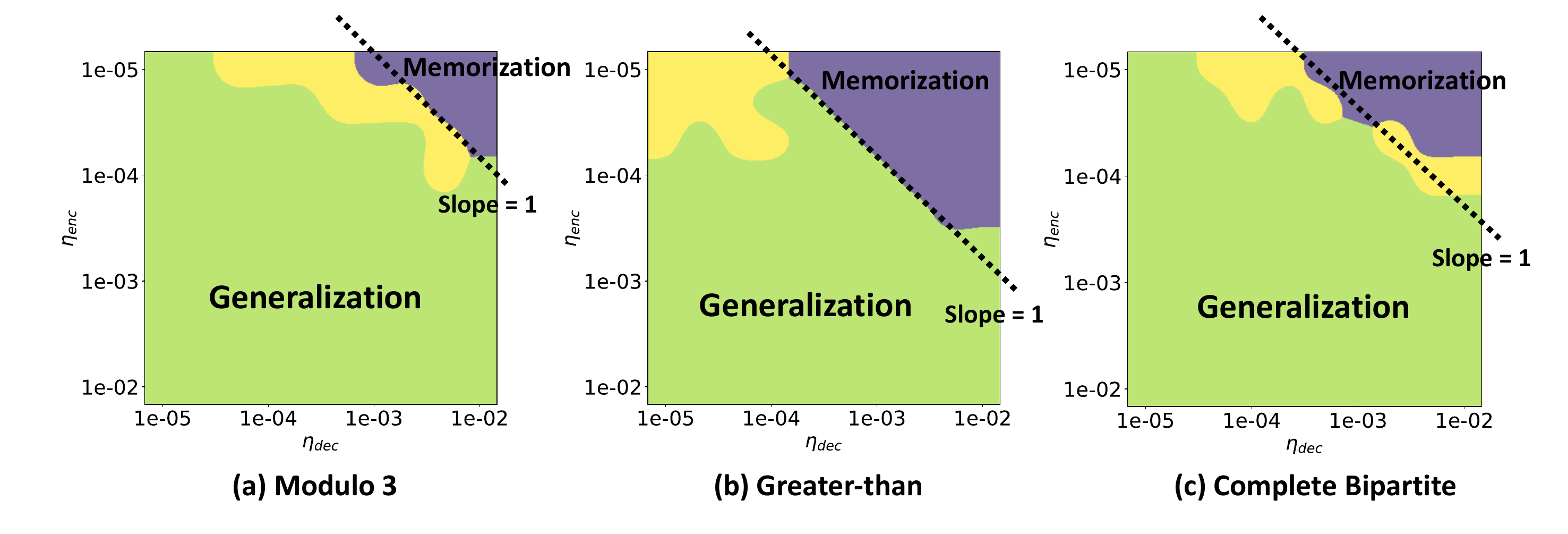}
\end{center}
\caption{Phase diagram as a function of the encoder learning rate $\eta_{\textrm{enc}}$ and decoder learning rate $\eta_{\textrm{dec}}$ for (a) modulo 3, (b) greater-than, and (c) complete bipartite relations. All figures indicate a slope-1 boundary between the generalization and memorization phases, as predicted by the interacting repon theory. Green, yellow, and purple regions indicate the generalization, delayed generalization (grokking), and memorization phase respectively.}
\label{repon-pd}
\end{figure*}

\noindent \textbf{(Theorem 3)} The probability of generalizable representation learning is upper-bounded by arctan formula Eq. (\ref{eq:prob}), which depends only on the ratio between learning rates.

\noindent \emph{Proof}: Let $p_r$ denote the probability of generalizable representation learning. According to Theorem 2, generalization occurs when the initial conditions lie within the green area of the phase space shown in Figure \ref{fig:repon-ps}. When the initial conditions $a_2^{(0)}$ and $c^{(0)}$ are drawn from a normal distribution with width $\sigma_a$ and $\sigma_c$ respectively, the probability of representation learning (\ie, $c^*\to 0$) is
\begin{equation}\label{eq:prob}
\begin{aligned}
    p_r&=\underset{\frac{c}{a}<\sqrt{\frac{\eta_x}{2\eta_A}}}{\iint}\frac{1}{2\pi\sigma_a\sigma_x}{\rm exp}\left(-\frac{1}{2}\left[\left(\frac{a}{\sigma_a}\right)^2+\left(\frac{c}{\sigma_c}\right)^2\right]\right)\mathrm{d}a\mathrm{d}c\\
    &=\frac{2}{\pi}{\rm arctan}\left(\frac{\sigma_a}{\sigma_c}\sqrt{\frac{\eta_x}{2\eta_A}}\right),
\end{aligned}
\end{equation}
which provides a practical prescription for enhancing representation learning. Note that this formula serves as an upper bound, since the repon collision is a necessary condition for generalization. Specifically, the analysis in Theorem 2 begins by examining the behavior when two repons come close to each other; however, they may never do so, depending on the initialization. Therefore, our final formula serves as an upper bound for the probability of representation learning. Below, we provide empirical evidence that supports this theory.

\begin{figure}
    \centering
    \includegraphics[width=0.8\linewidth]{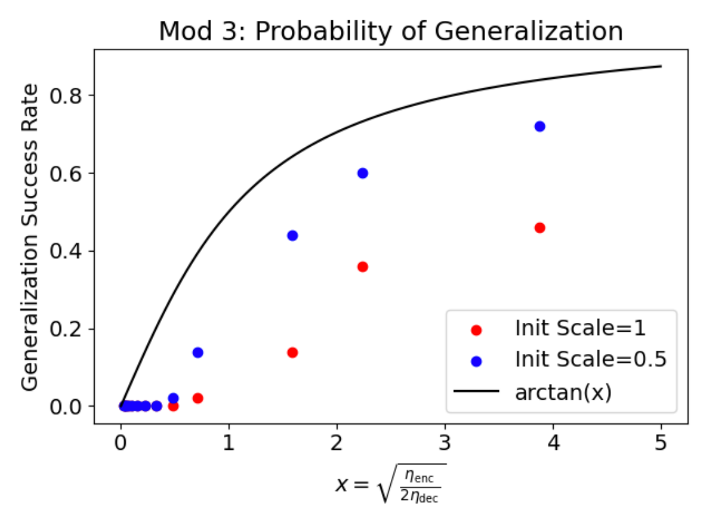}
    \caption{Probability of generalization as a function of the ratio between learning rates, for different initialization scales. \textcolor{black}{The initializations are drawn from a standard normal distribution scaled by the initialization factor.} As discussed in the main text, the probability of generalizable representation learning is upper bounded by Eq. (\ref{eq:prob}), which is plotted as the black solid line in the figure.}
    \label{fig:repon-prob}
\end{figure}

\textbf{Neural Network Experiments:} 
Figure \ref{repon-pd} shows the generalization/memorization phase diagram as a function of the encoder learning rate $\eta_{\textrm{enc}}$ and the decoder learning rate $\eta_{\textrm{dec}}$ for different relations. Phase classification criteria are described in Appendix \ref{app:def_phase}. Generalization can be interpreted as the probability of representation learning in Eq.(\ref{eq:prob}) exceeding a certain threshold. Hence, in a phase diagram as a function of encoder and decoder learning rates with logarithmic axes, one would expect the generalization-memorization boundary to have slope $-1$, which is indeed observed in Figure \ref{repon-pd}. Moreover, the phase boundary appears approximately at the same intercept, which also matches the prediction of Eq.~(\ref{eq:prob}) that the probability of representation learning should be task-independent, and only depends on the weight initialization. As a rule of thumb, we see that generalization requires that the learning rate ratio
$\eta_{\textrm{dec}} /\eta_{\textrm{enc}}\simlt 10$. This is consistent with the findings in \cite{liu2022towards}
that the 
ideal learning phase (comprehension) requires representation learning to be faster --- but not too much faster --- than the decoder.
We provide additional phase diagrams in Appendix \ref{app:addpds} that further validate the interacting repon theory. Figure \ref{fig:repon-prob} further supports the effective theory, indicating that the probability of generalizable representation learning is indeed upper bounded by Eq. (\ref{eq:prob}), which is shown as the black solid line in the figure.

\textbf{Discussions:} Our analysis is generalizable to multi-repon system. For instance, one could obtain an effective probability of representation learning for multi-repon system. For this, we consider a system consisting of one repon, and $N$ other repons clustered at the same position. By solving  similar equations of motion, one obtains the following modified probability of representation learning, which reduces to the result of Theorem 3 when $N=1$.
\begin{equation}
   p_r=\frac{2}{\pi}{\rm arctan}\left(\frac{\sigma_a}{\sigma_c}\sqrt{\frac{\eta_x}{\frac{4}{1+1/N}\eta_A}}\right) 
\end{equation}

Note that the process of deriving this corrected formula is very similar to how we solve many-body systems in physics, for example by defining quantities such as effective mass or effective force.

\section{Conclusion}
\label{conclusion}
We have presented an effective theory framework for understanding the statics and dynamics of model generalization in practice. In particular, we have studied the minimal data size needed, and critical decoder learning rates for generalization. While some static properties could be quite well described by our effective theory, incorporating more information about the input data's structures into the theory may contribute to further bridging the theory-experiment gap. There is ultimately a trade-off between the complexity of the theory and its accuracy. In this regard, we believe our information-theory-based approximation is helpful for understanding static properties of generalization. The theory of dynamic properties for generalization was largely well-supported by the relevant empirical phase diagrams. 

 Our study suggests the following unifying framework for understanding generalization: static properties of generalization can be derived from information-theoretic properties of the input data, while the dynamic properties of generalization can be understood by studying the system of interacting particles, repons. Our work also highlights the potential of physics-inspired effective theories for bridging the gap between theoretical predictions and practice in machine learning. \textcolor{black}{
We leave the empirical validation on more complex datasets, along with any necessary modifications to our framework, for future work.}

\begin{acknowledgments}
We thank the Center for Brains, Minds, and Machines (CBMM) for hospitality. This work was supported by The Casey and Family Foundation, the Foundational Questions Institute, the Rothberg Family Fund for Cognitive Science and IAIFI through NSF grant PHY-2019786.
\end{acknowledgments}

\appendix

\section{List of Default Hyperparameter Values}\label{app:def-param}
The hyperparameter values in Table \ref{tab:hyp} were used for the experiments unless specified otherwise.

\begin{table}[htbp]
\caption{List of default hyperparameter values.}
\begin{center}
	\begin{tabular}{c|c} \hline \hline
     Parameter & Default Value \\ \hline
    \centered{Embedding Space Dimension} & 2 \\ \hline 
    Weight Decay & 0 \\  \hline 
    Learning Rate & 0.001 \\  \hline
    Training Data Fraction & 0.75 \\  \hline 
    Decoder Depth & 3 \\  \hline 
    Decoder Width & 50 \\  \hline 
    Activation Function & Tanh \\  \hline 
        \hline
    \end{tabular}
    
        \label{tab:hyp}
    \end{center}
\end{table}

\section{Definition of Phases}\label{app:def_phase}

We have followed the phase classification criteria of \cite{liu2022towards}, which is shown in Table \ref{tab:phase-criteria}.
\begin{table*}[htbp]

\caption{Phase classification criteria.}
\begin{center}
\begin{tabular}{c|c|c|c}  
\hline \hline 
& \multicolumn{3}{c}{Criteria} \\ \hline 
Phase & train acc $> 0.9$ & test acc $> 0.9$ & (steps to test acc $> 0.9$) \\ 
& within $10^5$ steps & within $10^5$ steps & $-$ (steps to train acc $> 0.9$) $<$ $10^3$ \\ \hline 
Generalization & Yes & Yes & Yes \\    
Grokking & Yes & Yes & No \\ 
Memorization & Yes & No & N/A \\ \
Confusion & No & No & N/A \\ \hline \hline
\end{tabular}

\label{tab:phase-criteria}
\end{center}
\end{table*}

\section{Derivation of Eq.~(\ref{eq:gradient_flow})}\label{app:derivation_3}

The gradient flow is:
\begin{equation}
    \begin{aligned}
    \frac{d\mathbf{A}}{dt} &= - \eta_A\frac{\partial\ell}{\partial\mathbf{A}} \\
    &= -\eta_A [\mathbf{A}(\mathbf{x}_1\mathbf{x}_1^T+\mathbf{x}_2\mathbf{x}_2^T)+(\mathbf{b}-\mathbf{y})(\mathbf{x}_1+\mathbf{x}_2)^T], \\
    \frac{d\mathbf{b}}{dt} &= -\eta_A\frac{\partial\ell}{\partial\mathbf{b}} =  -\eta_A(\mathbf{A}(\mathbf{x}_1+\mathbf{x}_2)+\mathbf{b}-\mathbf{y}) \\
    \frac{d\mathbf{x}_i}{dt}&=-\eta_x\frac{\partial\ell}{\partial\mathbf{r}}=-\eta_x \mathbf{A}^T(\mathbf{A}\mathbf{x}_i+\mathbf{b}-\mathbf{y}),\ (i=1,2).
    \end{aligned}
\end{equation}
Without loss of generality, we can choose the coordinate system such that $\mathbf{x}_1+\mathbf{x}_2=\mathbf{0}$. For the $d\mathbf{b}/dt$ equation: setting the initial condition $\mathbf{b}(0)=\mathbf{0}$ gives $\mathbf{b}(t) = (1-e^{-2\eta_At})\mathbf{y}$. Inserting this into the first and third equations gives
\begin{equation}
\begin{aligned}
    &\frac{d\mathbf{A}}{dt} = -2\eta_A\mathbf{A}\mathbf{r}\mathbf{r}^T\\
    &\frac{d\mathbf{x}_i}{dt} = - \eta_x\mathbf{A}^T\mathbf{A}\mathbf{x}_i + \eta_xe^{-2\eta_At}\mathbf{A}^T\mathbf{y}
\end{aligned}
\end{equation}
Now define the relative position $\mathbf{r}\equiv(\mathbf{x}_1-\mathbf{x}_2)/2$ and subtracting $d\mathbf{x}_1/dt$ by $d\mathbf{x}_2/dt$ we get:
\begin{equation}
    \begin{aligned}
    &\frac{d\mathbf{A}}{dt} = -2\eta_A\mathbf{A}\mathbf{r}\mathbf{r}^T, \\
    &\frac{d\mathbf{r}}{dt} = - \eta_x\mathbf{A}^T\mathbf{A}\mathbf{r}. \\
    \end{aligned}
\end{equation}

\section{Monte Carlo Simulation Details}
\label{app:mc-details}
\textcolor{black}{\textbf{Equivalence Classes} We begin with the reference relation matrix $R_{\textrm{ref}}$ of equal modulo $n$. We then sample an arbitrary fraction of this matrix, based on the training fraction, and fill in all the inferrable entries according to the relation’s properties (in this case, symmetry, transitivity, and reflexivity). Using this information, we construct a dictionary where $pos[i]$ represents the set of classes to which $i$ can belong. If there exists j such that $R[i][j]=1$, then $pos[i]=[j\; \textrm{mod} \;n]$. If not, $pos[i]$ is set to the complement of the indices $j$ where $R[i][j]=0$}:

\[
\text{pos}[i] = \left\{
\begin{array}{ll}
[j\; \textrm{mod} \;n] & \text{if }  \exists j\; R[i][j] = 1, \\
\{j\; \textrm{mod} \;n \mid R[i][j] = 0\}^c & \text{otherwise}.
\end{array}
\right.
\]

\textcolor{black}{Next, we consider the problem of filling in all the entries $R[i][j]$ that remain undetermined. Assuming that each number is equally likely to belong to one of its candidate classes, we compute the probability $p_u$ that $R[i][j] = u$ ($u=0$ or $1$). The upper bound of prediction accuracy is then given by the sum of the number of determined entries and $\sum \text{max}(p_u,1-p_u)$ for all undetermined entries. We averaged the results over 20 different samplings. We applied the same methodology to the bipartite relation, since a complete bipartite relation is isomorphic to an equivalence relation with two classes.}

\textcolor{black}{\textbf{Total Orderings} We begin with the reference relation matrix $R_{\textrm{ref}}$ of the total ordering (greater-than relation). We then sample an arbitrary fraction of this matrix, based on the training fraction, and fill in all the inferrable entries according to the relation’s properties (in this case, antisymmetry and transitivity). Then, we randomly generate $1000$ total ordering sequences that are compatible with sampled entries of the relation matrix. Using these 10 sequences, we could compute the probability $p_u$ that $R[i][j] = u$ ($u=0$ or $1$). The remaining procedure is the same as for equivalence classes. Here, we averaged the results over 3 samplings to make the computations feasible, as each sampling requires generating 1000 sequences.}

\section{Optimal Model Complexity for Generalization}
\label{optimal_ma}

\cite{liu2022towards} postulated that encoder-decoder architectures have a \emph{Goldilocks zone} of decoder complexity such that generalization fails both when the decoder is too simple and when it is too complex. 
In this section, we validate this conclusion with our graph learning experiments.
Let the decoder have $n_h$ layers of uniform width $w$ so that the total number of trainable weight and bias parameters is approximately $w^2 n_h$. 

\begin{figure}[tb]
\begin{center}
\includegraphics[width=.7\linewidth]{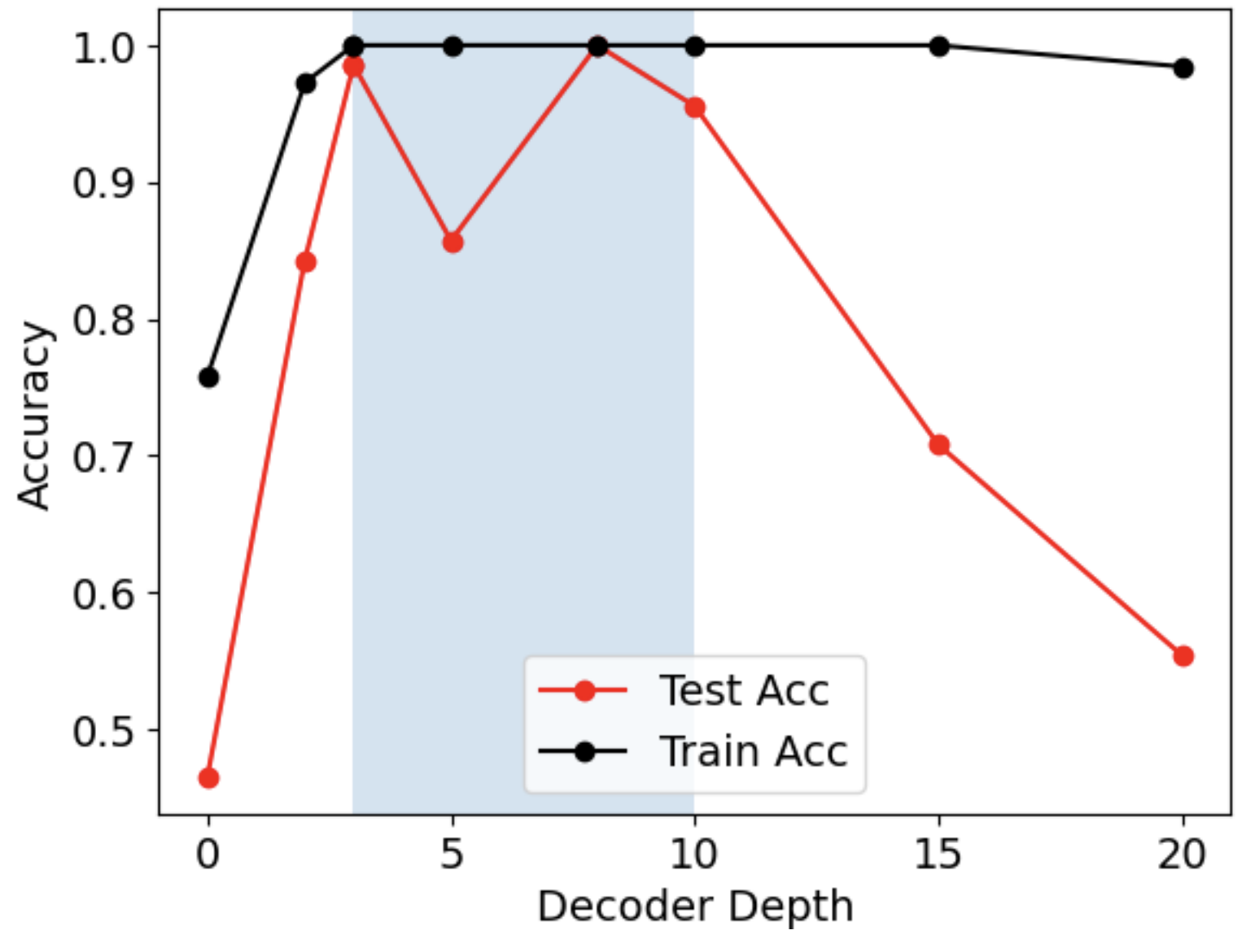}
\end{center}
\caption{Plot of test accuracy vs. decoder depth for learning Equivalent Modulo 3 relation, indicating the Goldilocks zone explained in the text. Decoder width $w=10$, and the training fraction of $0.3$ were used in the experiment.}
\label{goldilocks-fig}
\end{figure}

Figure \ref{goldilocks-fig} shows testing accuracy vs. decoder depth for \textcolor{black}{Modulo 3 relation. The plot} reveals the existence of a (i.e. sweet spot) where generalization occurs, degrading on both sides.
When the decoder is too simple, it lacks the ability to correctly decode learned embeddings, thus failing to provide any useful gradient information for the encoder --- the result is confusion, with not even the training accuracy becoming good.
When the decoder is too complex, it is able to correctly decode even random and extremely inconvenient encodings, again failing to 
provide any useful gradient information for the encoder --- the result is memorization (overfitting), with high training accuracy but low test accuracy. \textcolor{black}{In future work, we plan to develop an accompanying effective theory that explains this phenomenon.}

\section{Additional Phase Diagrams}\label{app:addpds}
\begin{figure*}[hp]
\begin{center}
\includegraphics[width=.95\linewidth]{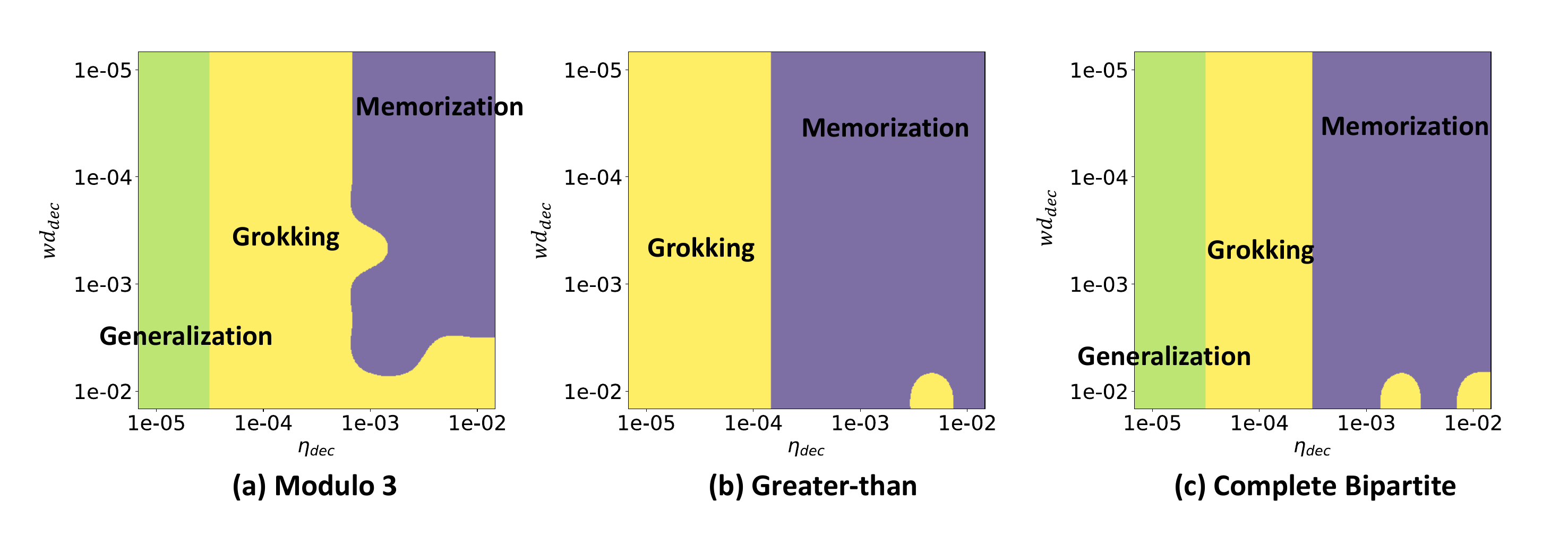}
\end{center}
\caption{Phase diagram as a function of the decoder's weight decay $wd_{\textrm{dec}}$ and the decoder's learning rate $\eta_{\textrm{dec}}$ for (a) modulo 3, (b) greater-than, and (c) complete bipartite relations. All figures indicate that the memorization-generalization boundary is independent of the weight decay, which is consistent with the result of interacting repon theory in Eq. (\ref{eq:prob}). Note that grokking is also a form of generalization, but the delayed generalization. Encoder's learning rate was fixed to $\eta_{\textrm{enc}}=10^{-5}$ in the experiment.}
\label{repon-pd-wd}
\end{figure*}

We provide additional phase diagrams in Figure \ref{repon-pd-wd}.

\bibliography{apssamp}

\end{document}